\documentclass[11pt,a4paper]{article}

\usepackage[T1]{fontenc}
\usepackage[utf8]{inputenc}
\usepackage{lmodern}
\usepackage{microtype}
\usepackage[a4paper,margin=28mm]{geometry}
\usepackage{authblk}
\usepackage{xurl}
\usepackage[
  backend=biber,
  style=numeric-comp,
  sorting=none,
  defernumbers=true,
  maxbibnames=99,
  giveninits=false,
  useprefix=true,
  doi=true,
  eprint=true,
  url=false,
  isbn=false
]{biblatex}
\DeclareFieldFormat[article,inproceedings,online]{title}{%
  \iffieldundef{url}{%
    \iffieldundef{doi}{\mkbibquote{#1}}{\href{https://doi.org/\thefield{doi}}{\mkbibquote{#1}}}%
  }{\href{\thefield{url}}{\mkbibquote{#1}}}%
}
\DeclareNameAlias{default}{family-given}
\DeclareNameAlias{sortname}{family-given}
\renewbibmacro*{url+urldate}{}
\usepackage[unicode=true,pdfencoding=auto,psdextra,hidelinks]{hyperref}
\hypersetup{
  pdfauthor={Giuseppe Marra; Pierre Geneves; Meghyn Bienvenu; Thomas Schiex; Sylvie Thiebaux; Silviu Maniu},
  pdftitle={Challenges, Opportunities, and Misconceptions in Neurosymbolic AI},
  pdfsubject={Perspectives from the 2026 CNRS Workshop Neurosymbolic Methods and Their Applications in Data Science}
}

\title{Challenges, Opportunities, and Misconceptions in Neurosymbolic AI\\[0.6em]
  {\large\normalfont\itshape
  Perspectives from the 2026 CNRS Workshop on Neurosymbolic Methods}}

\author[1]{Giuseppe Marra}
\author[2]{Pierre Genev\`es}
\author[3]{Meghyn Bienvenu}
\author[4]{Thomas Schiex}
\author[5,6]{Sylvie Thi\'ebaux}
\author[7]{Silviu Maniu}

\affil[1]{KU Leuven, Belgium}
\affil[2]{Univ. Grenoble Alpes, CNRS, Inria, Grenoble INP, LIG, Grenoble, France}
\affil[3]{Univ. Bordeaux, CNRS, Bordeaux INP, LaBRI, UMR 5800, Talence, France}
\affil[4]{MIAT, Universit\'e de Toulouse, ANITI, INRAE, Toulouse, France}
\affil[5]{LAAS-CNRS, Toulouse, France}
\affil[6]{Australian National University, Canberra, Australia}
\affil[7]{Univ. Grenoble Alpes, CNRS, Grenoble INP, LIG, France}

\date{}

\begin{document}
\maketitle

\begin{abstract}
We examine opportunities for neurosymbolic AI
alongside increasingly powerful neural models,
particularly with regards to efficiency, safety and
interpretability. We identify two major challenges:
ensuring semantic alignment between neural and
symbolic components, and managing the computational
cost of symbolic inference and learning. 
We discuss representative methodological approaches
to these challenges and address several common
misconceptions concerning differentiability,
independence assumptions, and scalability.
This paper synthesizes perspectives developed through discussions in the
\emph{Neurosymbolic Programming and Reasoning (NPR)}
working group at the 2026 CNRS workshop
\emph{Neurosymbolic Methods and Their Applications
in Data Science}, rather than providing an exhaustive survey.
\end{abstract}

\section{Introduction}\label{sec:intro}

Symbolic methods are already central to the development of the current generation of frontier
models. Tools extend the capabilities of LLMs with structured search, combinatorial solvers,
theorem provers and proof checkers both at inference time\footnote{Throughout this paper, we distinguish between
\emph{inference time}, referring to the use of a trained
model rather than training, and \emph{symbolic or
probabilistic inference}, referring to reasoning
computations such as marginalization, MAP optimization,
or model counting.}, and as sources of supervision for training or
fine-tuning, enabling them to solve problems beyond their standalone capabilities. This raises a
broader question: \textbf{where else can} \textbf{neurosymbolic AI add value alongside powerful
neural models?} We identify particularly promising opportunities in efficiency, constraints,
certification, and meaningful explanations.

We argue that academic research should develop foundational ideas while remaining motivated by practical problems. Foundational advances can themselves open up entirely new applications, some of which may be difficult to anticipate in advance. Competing directly with the industry to produce the next ``killer app''  looks difficult in most cases, because of the disproportionate amount of data (including copyrighted data) and compute used by current private actors in the field. However, credible application examples remain important: they can serve as demanding testbeds that expose limitations, motivate new theoretical developments, and demonstrate where neurosymbolic methods provide capabilities that purely neural approaches do not. In other words, theory and applications should inform each other.

A recurring tension in neurosymbolic AI (NeSy) is the balance between prior knowledge and data.
Symbolic knowledge, for instance in the form of constraints, can inject prior structure and
enforce desirable properties such as safety. Yet satisfying constraints alone does not guarantee
a useful design: incorrect or overly restrictive constraints may degrade performance or force
solutions away from the data distribution. The central question is therefore not simply whether
to impose symbolic structure, but what knowledge to encode, how strongly to enforce it, and
under which assumptions.

\section{Opportunities}\label{sec:opportunities}

\subsection{Effective AI Methods under Resource Limitations: Beyond Accuracy}\label{sec:resources}

The growing use of LLMs or foundation models for tasks that simpler tools could handle should be
reconsidered. Financial cost, environmental impact, computational resources, and training
requirements are important evaluation dimensions alongside accuracy.

One research hypothesis for NeSy is that explicitly encoding logical knowledge could reduce what
a neural model must learn, enabling smaller models, requiring less training data and providing better accuracy and generalizability. In
particular, symbolic knowledge encoded as constraints can reduce the hypothesis space and
thereby enable more compute and data efficient learning. This can be useful in applications that do not require
full foundation models, or for models distilled from larger ones.

Symbolic reasoning can also contribute capabilities beyond human imitation. Combinatorial solvers, such as SAT solvers, 
can solve problems requiring reasoning on millions of variables that no human can achieve. Hence, grafting this capacity 
directly inside neural models could grant them the ability to solve problems that are of higher complexity than what humans 
can achieve.

\subsection{Safety and Certification}\label{sec:safety}

\paragraph{Constraining agents during training and operation (``\emph{shielding}'').}

There is a clear distinction between encouraging compliance and enforcing it. Rewards and loss
penalties can guide a model toward desired behavior, but they permit trade-offs and do not by
themselves guarantee compliance on future inputs. An explicit decision layer can instead check,
restrict, repair, or reject proposed outputs and actions according to a formal specification.

Agent \emph{shielding} is a concrete example \cite{alshiekh2018shielding}. A shield can intervene during
operation and also during training, where it can prevent unsafe exploration. This is
particularly important when trial and error has physical or social consequences. Symbolic world
models and simulated environments may also support learning without requiring every interaction
to occur in reality. The issue does not apply only to embodied systems: software actions and influencing humans can also affect the world.

The guarantees provided by a shield depend on assumptions about perception and the environment.
Accounting for uncertainty may therefore require probabilistic guarantees, while reasoning over
longer horizons can substantially increase computational complexity.

\paragraph{Certification.}\label{sec:certification}

Certification offers a related opportunity to support safety and regulatory compliance.
Evaluating a model once is different from checking or enforcing requirements whenever it
operates. Training-time constraints can guide a model toward compliant behavior, but soft
penalties alone do not guarantee that future outputs will satisfy them. At inference time, a
symbolic layer can check or enforce constraints on outputs and actions, providing guarantees
relative to a formal specification.

The computational requirements of training-time guidance and inference-time certification may
differ substantially. Training can rely on relatively inexpensive approximate signals, whereas
inference-time certification may require stronger guarantees and therefore substancially more
computation. The difficulty depends on the underlying problem, but symbolic structure can
sometimes make certification tractable. When full certification is infeasible, alternatives
include probability bounds, certification of individual components, or conditional guarantees
within specified operating conditions.

Related approaches exist in programming languages and formal verification, for example to
enforce syntactic constraints on generated outputs or more generally formally specified
properties \cite{nagy2026chopchop}. A broader research challenge is to formalize the 
interactions among neural models, symbolic tools, and surrounding software, and 
to establish when guarantees on individual components extend to the behavior of the complete system.

\subsection{Out-of-Distribution Generalization}\label{sec:ood}
  
Compositionality and modularity suggest a research direction for generalization beyond the
training distribution. The hypothesis is that explicit representations of rules, relations, and
reusable components could help systems handle unfamiliar combinations of familiar elements.

Some encouraging results have been obtained in structured settings. Neural Logic Machines have demonstrated generalization from small training instances to larger ones in relational and algorithmic tasks, including sorting and graph reasoning \cite{dong2019nlm}. Neural-symbolic joint training has also demonstrated improved out-of-distribution generalization over neural baselines, from shorter to longer digit sequences in multi-digit addition and from fewer to more disks in optimal-action prediction for the Tower of Hanoi \cite{liu2023ood}.

More generally, research should test when such structure improves generalization and robustness, and how these
benefits depend on the validity of the encoded knowledge in the new setting.

\subsection{Interpretability and Explanations for Neurosymbolic Models}\label{sec:interpretability}

When the complexity of symbolic reasoning is bounded and the semantics of the neural and
symbolic components are aligned, neurosymbolic models can typically be regarded as interpretable
by design. In architectures based on neural predicates, these atomic symbols act as an
interpretable bottleneck, similarly to concept bottleneck models (CBMs) \cite{koh2020concept}. This
makes it possible to trace \emph{how} each prediction has been produced from \emph{which}
symbols. It also enables human users to act on these traces (for example, by correcting errors)
to improve predictions and potentially fine-tune the model to avoid repeating the same mistakes
on the same or similar instances.

Explanations are valuable for diagnosing failures, improving systems, and understanding when
reliance is justified. In settings where a model is already sufficiently trusted, however,
explanations may be an optional capability rather than a necessity.

One advantage of keeping the symbolic component explicit, rather than collapsing symbolic
knowledge into a sub-symbolic training objective (e.g., through semantic loss \cite{xu2018semantic} or
logic-based regularization), is that the symbolic reasoning process remains accessible at
inference time. This allows one to inspect how predicted symbols relate to the symbolic
constraints and how they contribute to the final prediction. Such explanations are faithful to
the explicit symbolic computation, but do not by themselves explain how the neural component
produced its symbols. Complementary approaches can be used to investigate this neural
computation, for instance by identifying computational circuits or by training replacement
models from which attribution graphs can be derived, as in \cite{conmy2023circuit} and \cite{ameisen2025circuit}.

\section{Challenges for the Application of Neurosymbolic AI}\label{sec:challenges}

\subsection{Shared Meaning Between Neural and Symbolic Components}\label{sec:alignment}

Combining neural models with symbolic reasoning requires a usable interface. Outputs of neural
models must be expressed in a language that the symbolic component can process. These are often explicit 
representations of probability or belief distributions over discrete variables (e.g., as neural predicates). 
But relevant symbolic properties of the inputs may often guide the design of neural architectures, 
as in geometric deep learning \cite{bronstein2021geometric}. More fundamentally,
\textbf{the model and the human-provided specification must attach compatible meanings to the
symbols and their properties.}

For example, suppose a neural network receives $0.7$ and $0.2$ as probabilities. Their meaning
imposes rules on how they should be combined: if they describe independent events, for instance,
the probability of both occurring is their product. However, a generic neural network can learn a mapping
that does not respect this probabilistic algebra. Simply constraining its output to lie between
$0$ and $1$ ensures a valid numerical range, but does not ensure that the computation preserves the
intended meaning of the inputs.
  
This semantic alignment is fundamental to providing meaningful guarantees about neurosymbolic
models. Such guarantees apply to the user's intended requirements only if the formal
specification faithfully captures that requirement and the model's computations respect the
specified meaning of the symbols. Otherwise, a model may provably satisfy a formal specification
while failing to deliver the behaviour the user intended.

The gap between optimal solution and intended meaning is the root cause of \emph{reasoning
shortcuts}~\cite{marconato2023shortcuts}: a model may satisfy symbolic constraints while exploiting
correlations or assigning unintended meanings to intermediate symbols.
An example discussed in~\cite{marconato2023shortcuts} involves a driving system that must stop at red lights or for pedestrians. The model may learn to represent pedestrians as red lights, since both lead to the same decision. This shortcut becomes problematic when the learned concepts are reused for an emergency vehicle that may cross red lights but must still stop for pedestrians. 
A correct final answer therefore does not, by itself, establish that the system used the intended reasoning process or
interpreted the symbols consistently with the user's specification. To further illustrate the issue, consider a very weakly supervised variant of the MNIST addition benchmark used in DeepProbLog \cite{manhaeve2018deepproblog}. 
Given three digit images, suppose that the only supervision is
the constraint that the predicted digits for the first two images must sum to the predicted digit for the third (and not to a known number, as in the original task). This objective
admits a trivial solution: assigning 0 to every image, since $0 + 0 = 0$. The model can therefore satisfy the symbolic rule
while violating the user's intended interpretation. In essence, this is an identifiability issue: the weak supervision of the learning task admits naive solutions (here symbol groundings) that the task designer did not envision.

\subsection{Cost of Training and Evaluation}\label{sec:cost}

Computational cost remains a challenge in neurosymbolic AI. In probabilistic approaches,
training can repeatedly require inference over a combinatorial space of symbolic assignments;
exact weighted model counting is \#P-hard \cite{dubray2024anytime}. Compiling knowledge into arithmetic
circuits makes evaluation efficient relative to circuit size but may lead to an exponential blow-up 
in circuit size: compilation, memory use and repeated evaluation can still be costly or impossible \cite{maene2025klay}. 
Recent approaches also seek to avoid full provenance materialization \cite{valentin2026dpnl}, 
although the resulting search can itself remain costly.
 
Computational difficulty can arise from the underlying problem itself, for example when
reasoning involves NP-hard combinatorial problems. Exact probabilistic certification is also
particularly demanding. However, this should not become a blanket claim that symbolic methods
cannot scale. (Provably) approximate methods and heuristics should still be investigated for practical
applications.

\section{Illustrative Applications}\label{sec:applications}
 
Several application domains provide motivating test cases for neurosymbolic methods. 
Robotics, autonomous driving, planning, and code generation provide natural settings in which learned models must interact with explicit constraints, structured reasoning, or verification mechanisms. 

Protein design provides a concrete example. Desired proteins may need to satisfy combinations of
constraints on amino-acid counts, charge, particular positions, or the number of distinct amino
acids used. Pure neural sampling can struggle to satisfy such combinations reliably, motivating
approaches that couple learned models with explicit constraint reasoning \cite{defresne2026generative}.

Recent AI systems also illustrate the broader combination of neural models with symbolic tools.
AlphaProof \cite{hubert2026olympiad}, as well as agentic systems combining LLMs with formal verifiers such
as Lean-based proof checkers \cite{yang2023leandojo}, are prominent examples of architectures in which
neural generation and symbolic reasoning or verification play complementary roles.

\section{Promising Methodologies and Their Current Limits}\label{sec:methodo}
 
The challenges identified above can be addressed through several complementary methodological
directions. The approaches below are illustrative rather than exhaustive, and each addresses
only part of the corresponding challenge.

\subsection{Methodologies for the Semantic Alignment Challenge}\label{sec:methodo-alignment}

Several approaches seek to reduce semantic misalignment by making intermediate concepts
explicit, constraining their relationships, or allowing human feedback to act directly on them.
 
\begin{itemize}
\item 
  \textbf{Concept-bottleneck models.} Concept bottleneck models supervise intermediate
  representations with human-defined concepts, making the correspondence between model variables
  and intended meanings explicit and inspectable \cite{koh2020concept}. Their effectiveness, however,
  depends on defining an appropriate concept vocabulary and typically requires direct concept-level
  supervision, in contrast  with many neuro-symbolic architectures advertised ability to train neural 
  models through weak supervision (as in the original MNIST addition task previously mentioned).
\item
  \textbf{Preference-based alignment.} Human comparisons can provide an indirect route toward
  aligning model behavior with human intent. In RLHF, for example, human preferences are turned
  into a reward signal that encourages behavior consistent with human judgments \cite{ouyang2022training}.
  Besides the lack of guarantees of these methods in a general setup, the feedback operates primarily at the level of behavior, however, and does not by itself ensure that intermediate symbols acquire their intended semantics.
\item
  \textbf{Label-free concept bottleneck models.} Label-free CBMs use pretrained language and
  vision-language models to generate and associate latent representations with named concepts
  without manually annotated concept labels \cite{oikarinen2023labelfree}. This reduces the supervision burden
  but may inherit both useful semantic knowledge and semantic mismatches from the pretrained
  models used to construct the concepts.
\item
  \textbf{Symbolic constraints as learning signals.} User-specified relationships among symbols
  can be incorporated as constraints during learning, narrowing the space of admissible
  interpretations \cite{xu2018semantic}. This provides a specifically neurosymbolic route to reducing
  semantic misalignment, but, as already said, satisfying the constraints does not necessarily identify the
  intended interpretation uniquely.
\item
  \textbf{Interactive concept correction.} Users can intervene on intermediate concept
  predictions and correct them when the model's interpretation diverges from theirs~\cite{shin2023intervention}. Such interventions provide feedback directly at the semantic interface, but
  require concepts that are understandable and actionable by human users.
\end{itemize}

\subsection{Methodologies for the Computational Cost Challenge}\label{sec:methodo-cost}

The computational cost of neurosymbolic systems has multiple sources and can therefore be
addressed through complementary strategies. Current approaches include approximating or
interrupting symbolic or probabilistic inference, accelerating its execution, and reducing how often expensive inference must be performed.

\begin{itemize}

\item
  \textbf{Approximate and anytime methods.} One direction is to \emph{learn from approximate
  inference without exhaustively evaluating the symbolic space.} EXAL constructs a
  sampling-based learning objective from sampled and reweighted explanations \cite{verreet2024explain},
  while IndeCateR provides unbiased gradient estimates with lower variance than REINFORCE for
  independent categorical variables \cite{desmet2023catlog}. Anytime algorithms offer another useful
  capability: weighted model counting can progressively refine certified lower and upper bounds,
  allowing inference to stop when the available computational budget is exhausted or the desired
  precision is reached \cite{dubray2024anytime}. Related search-based approaches can avoid full
  provenance materialization: DPNL \cite{valentin2026dpnl} uses oracle-guided DPLL-style lazy
  exploration, while ApproxDPNL adds early stopping with certified bounds on the exact output
  probability. These approaches trade exhaustive computation for search, or progressively
  refined bounds, and their efficiency ultimately depends on how effectively the structure of
  the problem can be exploited.
\item
  \textbf{Acceleration}. A second direction is to \emph{accelerate symbolic or probabilistic
  inference and better exploit modern hardware}. KLay reorganizes arithmetic circuits for
  parallel GPU evaluation and differentiation \cite{maene2025klay}. Pyro provides broader
  infrastructure for GPU-accelerated stochastic variational inference, supporting scalable
  approximation for compatible probabilistic models \cite{bingham2019pyro}. Such acceleration can
  substantially reduce execution costs, but does not remove the underlying combinatorial
  complexity or the cost of constructing large symbolic representations.
\item
  \textbf{Relaxations during training and amortized inference}. A third direction is to
  \emph{reduce the need for expensive exact} \emph{inference during training or repeated
  inference}. Relational Neural Machines use local likelihood approximations and relaxed MAP
  inference \cite{marra2020relational}. Pseudolikelihood-based training similarly replaces global
  normalisation with local conditional calculations, while retaining discrete solving with
  Toulbar2 for prediction, as demonstrated in neuro-symbolic combinatorial reasoning~\cite{defresne2023scalable,defresne2026scaling} 
  and protein design~\cite{defresne2026generative}. A-NeSI takes an amortized approach, training neural models to approximate probabilistic inference using symbolic simulations so that repeated probabilistic queries can subsequently be answered efficiently \cite{vankrieken2023anesi}. These
  approaches reduce computational cost by replacing or amortizing global inference, at the price
  of relying on approximations whose quality must be assessed for the task and distribution of
  interest.
\end{itemize}

\section{Misconceptions}\label{sec:misconceptions}

Several common misconceptions about neurosymbolic AI arise from properties of particular models
or inference procedures rather than from intrinsic limitations of the field.

\subsection{The symbolic component must be differentiable}

Training neural models by gradient descent does not require
the symbolic reasoning component itself to be differentiable.
The key distinction is between obtaining gradients with
respect to neural parameters and differentiating through
the symbolic reasoning procedure.

DPNL illustrates this distinction: it computes output
probabilities through DPLL-style oracle-guided inference.
These probabilities are differentiable with respect to
the neural predictions, even though the symbolic function
and its oracle need not be differentiable
\cite{valentin2026dpnl}.

EM-NeSy illustrates a complementary approach: posterior
inference over neurally predicted symbols provides supervision,
while gradients are propagated only through the neural
component \cite{seibt2026emnesy}.
Any exact or approximate inference engine that provides
the required posterior information can play this role,
regardless of its differentiability.
Under exact inference, EM-NeSy recovers the standard
end-to-end NeSy gradient; approximate inference instead
changes the learning signal and potentially its guarantees
\cite{seibt2026emnesy}.

Abductive learning follows another strategy: logical
abduction is used to infer or revise symbolic facts so
that they are consistent with the available observations
and rules, and these revised facts are then used as
supervision for the perception model \cite{dai2019abductive}.

Other approaches construct surrogate gradients for discrete
reasoning or optimization. I-MLE repeatedly calls a discrete
solver to find MAP solutions (that is,
the highest-scoring discrete assignments) under perturbations,
and uses the differences between these solutions to estimate
a learning signal \cite{niepert2021implicit}.
Black-box differentiation similarly treats a combinatorial
solver as a non-differentiable black box during the forward
computation, but constructs a surrogate gradient from
additional solver calls so that the neural component can
still be trained \cite{vlastelica2020blackbox}.

These methods enable gradient-based learning without requiring
the symbolic reasoning component itself to be differentiable.

\subsection{Symbolic concepts must be independent}

Independence is a modelling choice, not a general requirement of neurosymbolic AI. Graphical models, Markov logic
and their neural extensions (NMLN, RNM, NeuPSL, Effie and the E-PLL loss) represent joint distributions in which symbolic
variables can directly depend on one another \cite{richardson2006markov,marra2021neuralmarkov,marra2020relational,pryor2023neupsl,defresne2023scalable,defresne2026scaling}.

DeepProbLog uses independent primitive random choices, but its rules can express dependent
concepts through shared causes and conditional constructions \cite{manhaeve2018deepproblog}. Choosing separate,
independent neural predictions therefore restricts a particular model rather than the language's
expressiveness. Results on uncertainty and reasoning shortcuts under independence remain
relevant precisely when that factorisation is imposed \cite{vankrieken2025shortcuts}; they do not establish a
limitation of all NeSy models. Indeed, the E-PLL based architecture~\cite{defresne2023scalable,defresne2026scaling} can 
not only constrain independent concepts by logical rules or probabilistic knowledge but also learn these in the form of a 
pairwise graphical model~\cite{cooper2020graphical}.

\subsection{Probabilistic circuits and NeSy cannot scale}

Scalability depends on the representation, the inference task, the structure of the problem, and
the required accuracy. For suitable probabilistic circuits, many inference operations can be
performed in time linear in the size of the circuit, although compilation and circuit size can
themselves become bottlenecks \cite{maene2025klay}, as the worst-case size of the circuits grows exponentially with the input formula. 
As discussed above, GPU acceleration, sampling,
approximate or amortized inference, and lazy search address different sources of computational
cost \cite{maene2025klay,verreet2024explain,desmet2023catlog,marra2020relational,defresne2023scalable,vankrieken2023anesi,dubray2024anytime,bingham2019pyro,valentin2026dpnl}. 
With approximation such as pseudo-likelihoods, training can even provide fully scalable training, enabling
training on large datasets including samples of several thousands simultaneously interacting amino acids in protein design~\cite{defresne2026generative}.
These methods retain familiar trade-offs of probabilistic
modelling: exact inference can remain computationally hard, while approximation can affect
accuracy or guarantees. This calls for choosing suitable algorithms, especially when domain knowledge is available, and evaluating their
scaling, rather than treating intractable exact inference as evidence that NeSy as a whole
cannot scale.

\section*{Acknowledgments}\phantomsection\label{acknowledgments}
This document emerged from discussions within the
\emph{Neurosymbolic Programming and Reasoning (NPR)}
working group, co-chaired by Pierre Genev\`es and
Thomas Schiex, at the CNRS workshop
\emph{Neurosymbolic Methods and Their Applications
in Data Science} (October 6--9, 2026). 
The workshop received financial support from the
CNRS research groups (GDRs) MaDICS and RADIA,
and was organized in collaboration with the
Laboratoire d'Informatique de Grenoble (LIG),
the MIAI NSRL Chair, and LabEx PERSYVAL-Lab,
with the involvement of RECAST, the joint
working group of the two GDRs.

\section*{References}\phantomsection\label{refs}
\begingroup
\small
\printbibliography[heading=none,notkeyword=survey,resetnumbers=true]
\endgroup

\section*{Surveys}\phantomsection\label{surveys}
For broader surveys and taxonomies of neurosymbolic AI, we refer the reader to
\cite{marra2024survey,colelough2025systematic,vanbekkum2021modular,chaudhuri2021programming,delong2025kg,gibaut2023taxonomy,renkhoff2024verification}.

\begin{refcontext}[labelprefix=S]
\begingroup
\small
\printbibliography[heading=none,keyword=survey,resetnumbers=true]
\endgroup
\end{refcontext}

\end{document}